\documentclass[conference]{IEEEtran}
\IEEEoverridecommandlockouts

\usepackage{cite}
\usepackage{amsmath,amssymb,amsfonts}
\usepackage{algorithmic}
\usepackage{graphicx}
\usepackage{textcomp}
\usepackage{xcolor}
\usepackage{tikz}
\usetikzlibrary{arrows.meta,positioning}
\usepackage{booktabs}
\usepackage{placeins}
\usepackage{eso-pic}
\usepackage[caption=false,font=footnotesize]{subfig}

\usepackage{doi}

\usepackage[nameinlink,capitalise,noabbrev]{cleveref}
\crefname{figure}{Fig.}{Figs.}
\Crefname{figure}{Fig.}{Figs.}
\crefname{table}{Table}{Tables}
\Crefname{table}{Table}{Tables}
\crefname{equation}{Eq.}{Eqs.}
\Crefname{equation}{Eq.}{Eqs.}

\begin{document}

\AddToShipoutPictureFG*{%
  \AtPageUpperLeft{%
    \put(52,-23){%
      \parbox[t]{\dimexpr\paperwidth-104pt\relax}{%
        \fontsize{8}{9}\selectfont
        2026 29th International Conference on Computer
        and Information Technology (ICCIT)\\
        18--20 December 2026, Cox's Bazar, Bangladesh
      }%
    }%
  }%
}

\title{
CAFE-AQI: Coverage-Aware Day-Ahead AQI Forecasting with Early Warning and Conformal Uncertainty in Bangladesh
}

\author{
\IEEEauthorblockN{Md Rafid Islam}
\IEEEauthorblockA{
Electrical and Computer Engineering\\
North South University\\
Dhaka, Bangladesh\\
md.islam.241@northsouth.edu
}

\and

\IEEEauthorblockN{Rafsan Jany}
\IEEEauthorblockA{
Digital Health Research Division, KIOM\\
University of Science and Technology\\
Daejeon, Republic of Korea\\
rafsanjany@ust.ac.kr
}

\and

\IEEEauthorblockN{Rafid Bin Zaman}
\IEEEauthorblockA{
Civil and Environmental Engineering\\
Islamic University of Technology\\
Gazipur, Bangladesh\\
rafidbinzaman@iut-dhaka.edu
}

\and

\IEEEauthorblockN{Mohammad Fahim Islam}
\IEEEauthorblockA{
Computer Science and Engineering\\
American International University-Bangladesh\\
Dhaka, Bangladesh\\
23-51971-2@student.aiub.edu
}

\and

\IEEEauthorblockN{Hafiz Abdur Rahman\textsuperscript{*}}
\IEEEauthorblockA{
Electrical and Computer Engineering\\
North South University\\
Dhaka, Bangladesh\\
hafiz.rahman@northsouth.edu\\
\textsuperscript{*}Corresponding Author
}
}
\maketitle

\begin{abstract}
Next-day air quality forecasting often assumes city records are equally complete and independent. However, an audit of a large Bangladesh dataset revealed a sharp coverage expansion after August 2022 and showed that 103 city labels correspond to only 61 unique AQI trajectories. To address these issues, this paper presents CAFE-AQI, which is a coverage-aware framework for forecasting next-day maximum AQI. It also estimates the probability of Unhealthy-or-worse conditions (AQI $\ge$ 151). Data were divided chronologically into training (2022--2023), calibration (2024), and testing (2025). In order to prevent duplicate records from dominating training and evaluation, identical trajectories were weighted. The framework is a combination of temporal and pollutant features with a five-neighbor spatial context. It also includes a class-balanced warning head and risk-conditional conformal calibration with an aim to provide early warnings and prediction intervals. Evaluation on the duplicate-balanced 2025 data gave an MAE of 9.76, an RMSE of 15.04, and an \(R^2\) of 0.910 for CAFE-AQI. Its MAE was 3.63 AQI lower than persistence (95\% CI: 3.49--3.76), with lower errors in each of the 61 trajectory groups. For unhealthy-air warnings, the \(F_1\)-score was 0.908 and the AUROC was 0.982. The nominal 90\% intervals covered 92.09\% of observations. When an entire spatial block was excluded from training, the MAE remained 10.66, showing that the lightweight model transferred reasonably well to unseen locations.
\end{abstract}


\section{Introduction}

Air pollution is both a long-term health burden and a daily warning challenge. The WHO identifies particulate matter, O$_3$, NO$_2$, SO$_2$, and CO as major health threats \cite{who2021aqg}, while the World Bank has emphasized the need for stronger monitoring and early-warning systems in Bangladesh \cite{worldbank2022breathing}. A practical next-day system should therefore forecast AQI severity and warn when it may exceed 151, the start of the \emph{Unhealthy} category \cite{usepa2024aqi}.

Machine-learning models can identify nonlinear temporal and spatial patterns. Because of their efficacy, these models have replaced time-series methods in air quality forecasting \cite{yi2018deepair,agbehadji2024review}. However, accuracy can be misleading when the evaluation ignores data structure. Random temporal splits may leak future information, while separating city labels that share the same trajectory can create false spatial generalization. Reliable evaluation must preserve temporal order, and along with that, it should also account for dependent observations \cite{cerqueira2020evaluation,roberts2017crossvalidation}.

These issues are especially important for the Bangladesh AQI dataset used here. It contains 3,193,198 hourly records from 103 labeled cities. But geographic coverage expanded in August 2022. The city labels represent only 61 unique AQI trajectories. Treating all labels as independent could result in duplicate evidence across training and testing. It is also possible that a random split could mix past, future, and identical trajectories. We first audit these dependencies to define a valid forecasting period and evaluation setting.

We propose CAFE-AQI, which is a lightweight, coverage-aware framework that predicts the next-day maximum AQI and the probability of AQI reaching the Unhealthy threshold of 151. It combines causal temporal and pollutant features with information from five nearest neighbors. A dedicated warning head handles class imbalance along with risk-conditional split-conformal intervals that represent prediction uncertainty \cite{angelopoulos2023conformal}.

The main contributions are:

\begin{itemize}
    \item A coverage and duplication audit identifying the August 2022 data transition and 61 unique trajectory groups among 103 city labels.

    \item A lightweight dual-head framework for next-day AQI forecasting, Unhealthy-air warnings, uncertainty intervals, and grouped explanations.

    \item A leakage-resistant protocol using 2022--2023 for training, 2024 for calibration, and 2025 for final testing, supported by baselines, ablations, group-level uncertainty analysis, and unseen-group transfer tests.
\end{itemize}

\section{Data Audit and Problem Formulation}
\label{sec:data_audit}

\subsection{Data Product and Temporal Coverage}

The analyzed data product contains 3,193,198 hourly records associated with 103 Bangladeshi city labels and spans January 1, 2000, to November 23, 2025~\cite{paul2026aqi}. This dataset has also been used for the purpose of characterizing long-term air quality trends, seasonal patterns, and spatial pollution regimes across Bangladesh~\cite{islam2025twentyfive}. Each record provides a timestamp, city identifier, geographic coordinates, AQI, and seven atmospheric variables: PM$_{10}$, PM$_{2.5}$, CO, CO$_2$, NO$_2$, SO$_2$, and O$_3$. Because the available metadata do not establish that every row represents an observation from a verified physical monitoring station, we refer to these records as \emph{city-referenced air-quality estimates} rather than station measurements.

The dataset has data from 103 locations, but before August 4, 2022, only 2 of them actually had data. After that date, all 103 started reporting. Also, the way AQI and CO were measured changed on that same date. We therefore use only the common coverage period beginning August 4, 2022. Forecasting starts on August 19, after the required 14-day history.

The dataset had 3,193,198 records with no duplicates. 2,472 rows were removed because AQI was missing, and 3,190,726 valid records remained. CO$_2$ was dropped since most of its values (69.50\%) were missing. Hourly data was grouped into daily averages, and a day was only removed if the AQI or a necessary feature was missing.

\subsection{Duplicate-Trajectory Audit}

No duplicate rows were found, but this does not mean that all 103 city series are independent. We created a complete daily maximum-AQI series for each city, retained missing dates, and compared series with matching SHA-256 hashes.

The audit found only 61 unique trajectories. Twenty-eight groups contained two to four cities with identical AQI values and missingness patterns, while 33 were single cities. To prevent repeated trajectories from receiving extra influence, each city-day was weighted as
\begin{equation}
    w_{c,t}=\frac{1}{m_{g(c)}},
    \label{eq:trajectory_weight}
\end{equation}
where \(m_{g(c)}\) is the number of cities sharing the same trajectory. These weights were used for model fitting and evaluation. Cities with identical trajectories were also kept in the same partition during transfer experiments to prevent leakage.

\subsection{Forecasting Tasks and Leakage Control}

For city \(c\), CAFE-AQI predicts the maximum AQI of the following calendar day:
\begin{equation}
y_{c,t+1}=\max_{h\in\mathcal{H}(t+1)}\mathrm{AQI}_{c,h},
\qquad
z_{c,t+1}=\mathbb{I}(y_{c,t+1}\geq151),
\label{eq:targets}
\end{equation}
where \(z_{c,t+1}\) indicates an \emph{Unhealthy}-or-worse day under the U.S.\ AQI definition \cite{usepa2024aqi}. Using only information available by the end of day \(t\), the model estimates
\begin{equation}
\hat{y}_{c,t+1}=f_{\theta}(\mathbf{x}_{c,t}),
\qquad
\hat{p}_{c,t+1}=g_{\phi}(\mathbf{x}_{c,t}).
\label{eq:dual_task}
\end{equation}

Only consecutive day pairs are retained, and the split follows the target date. City identity is not used as a feature. After removing incomplete pairs, 122,879 city-days remain. Regression is trained with trajectory-group weights:
\begin{equation}
\min_{\theta}
\sum_{(c,t)\in\mathcal{D}_{\mathrm{train}}}
w_{c,t}\,
\ell\!\left(y_{c,t+1},f_{\theta}(\mathbf{x}_{c,t})\right),
\label{eq:weighted_objective}
\end{equation}
which prevents duplicated trajectories from receiving greater influence simply because they appear under multiple city labels.

\section{Proposed CAFE-AQI Framework}
\label{sec:framework}

\subsection{Framework Overview}

\begin{figure*}[!t]
\centering

\resizebox{0.90\textwidth}{!}{%
\begin{tikzpicture}[
    node distance=9mm and 9mm,
    block/.style={
        rectangle,
        rounded corners=3pt,
        draw=black!55,
        fill=gray!4,
        line width=0.8pt,
        minimum width=2.65cm,
        minimum height=1.20cm,
        text width=2.35cm,
        align=center,
        font=\small,
        inner sep=4pt
    },
    prep/.style={
        block,
        draw=teal!65!black,
        fill=teal!6,
        text width=2.75cm,
        minimum width=3.05cm
    },
    reg/.style={
        block,
        draw=blue!65!black,
        fill=blue!6
    },
    warn/.style={
        block,
        draw=orange!80!black,
        fill=orange!7
    },
    calib/.style={
        block,
        draw=purple!65!black,
        fill=purple!6
    },
    outputreg/.style={
        block,
        draw=blue!65!black,
        fill=blue!10,
        line width=1pt
    },
    outputwarn/.style={
        block,
        draw=orange!80!black,
        fill=orange!10,
        line width=1pt
    },
    explain/.style={
        block,
        draw=gray!65!black,
        fill=gray!7
    },
    arrow/.style={
        -{Latex[length=2.2mm,width=1.5mm]},
        line width=0.85pt,
        draw=black!55
    },
    regarrow/.style={
        arrow,
        draw=blue!65!black
    },
    warnarrow/.style={
        arrow,
        draw=orange!80!black
    },
    explainarrow/.style={
        arrow,
        draw=gray!60!black,
        densely dashed
    }
]

\node[block] (raw) {
    \textbf{Hourly AQI data}\\[1pt]
    {\scriptsize 3.19\,M records}\\[-1pt]
    {\scriptsize 103 city labels}
};

\node[prep, right=of raw] (preparation) {
    \textbf{Audit and preparation}\\[1pt]
    {\scriptsize August 2022 regime}\\[-1pt]
    {\scriptsize 61 trajectory groups}\\[-1pt]
    {\scriptsize causal city-day features}
};

\node[reg, right=13mm of preparation, yshift=10mm] (regressor) {
    \textbf{AQI regressor}\\[1pt]
    {\scriptsize 30 features}\\[-1pt]
    {\scriptsize trained: 2022--2023}
};

\node[calib, right=of regressor] (conformal) {
    \textbf{Conformal interval}\\[1pt]
    {\scriptsize nominal 90\%}\\[-1pt]
    {\scriptsize calibrated: 2024}
};

\node[outputreg, right=of conformal] (forecast) {
    \textbf{Day-ahead AQI}\\[1pt]
    {\scriptsize maximum forecast}\\[-1pt]
    {\scriptsize point + interval}\\[-1pt]
    {\scriptsize tested: 2025}
};

\node[explain, right=of forecast] (importance) {
    \textbf{Regression audit}\\[1pt]
    {\scriptsize grouped permutation}\\[-1pt]
    {\scriptsize importance}
};

\node[warn, right=13mm of preparation, yshift=-10mm] (classifier) {
    \textbf{Warning classifier}\\[1pt]
    {\scriptsize 24 features}\\[-1pt]
    {\scriptsize trained: 2022--2023}
};

\node[calib, right=of classifier] (threshold) {
    \textbf{Warning threshold}\\[1pt]
    {\scriptsize maximum-$F_1$}\\[-1pt]
    {\scriptsize calibrated: 2024}
};

\node[outputwarn, right=of threshold] (warning) {
    \textbf{Unhealthy-air alert}\\[1pt]
    {\scriptsize $P(\mathrm{AQI}_{t+1}\geq151)$}\\[-1pt]
    {\scriptsize probability + alert}\\[-1pt]
    {\scriptsize tested: 2025}
};

\node[explain, right=of warning] (occlusion) {
    \textbf{Warning audit}\\[1pt]
    {\scriptsize local feature-group}\\[-1pt]
    {\scriptsize occlusion}
};

\draw[arrow] (raw) -- (preparation);

\draw[regarrow]
    (preparation.east) -- ++(5mm,0) |- (regressor.west);

\draw[warnarrow]
    (preparation.east) -- ++(5mm,0) |- (classifier.west);

\draw[regarrow] (regressor) -- (conformal);
\draw[regarrow] (conformal) -- (forecast);
\draw[explainarrow] (forecast) -- (importance);

\draw[warnarrow] (classifier) -- (threshold);
\draw[warnarrow] (threshold) -- (warning);
\draw[explainarrow] (warning) -- (occlusion);

\end{tikzpicture}%
}

\vspace{-2mm}
\caption{CAFE-AQI workflow. The task-specific models are trained on 2022--2023 data, calibrated on 2024, and evaluated on the untouched 2025 test set.}
\label{fig:cafe_framework}

\end{figure*}
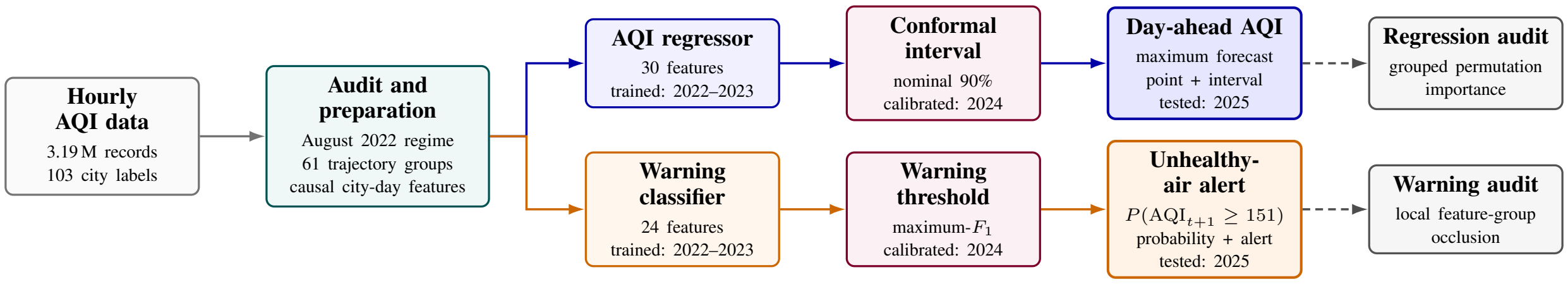

CAFE-AQI is designed as a lightweight forecasting and early-warning framework rather than a computationally heavy forecasting network. As illustrated in Fig.~\ref{fig:cafe_framework}, the audited city-day table is first converted into causal temporal, pollutant, and geographic features. We built two models: one to predict tomorrow's peak AQI (with a special confidence range) and another to predict the chance of unhealthy air. We gave equal importance to repeated weather patterns, ignored city names, and finalized all model settings before testing on 2025 data.

The term \emph{dual-head} refers here to two separate learners fitted on the same audited forecasting table. It does not imply a shared neural encoder. This separation allows the continuous forecast and the public-health warning to be optimized for their respective objectives.

\subsection{Causal Temporal and Neighbour Features}

For every city \(c\) and forecast date \(t\), the regression input is
\begin{equation}
    \mathbf{x}^{\mathrm{reg}}_{c,t}
    =
    \left[
        \mathbf{x}^{\mathrm{temp}}_{c,t},
        \mathbf{x}^{\mathrm{poll}}_{c,t},
        \mathbf{x}^{\mathrm{coord}}_{c},
        \mathbf{x}^{\mathrm{nei}}_{c,t}
    \right]
    \in\mathbb{R}^{30}.
    \label{eq:regression_features}
\end{equation}

The model uses 24 features total. Twelve of these features describe the current day's AQI, recent changes, and seasonal patterns. They include daily statistics (maximum, mean, standard deviation), lagged values (1 and 7 days ago), rolling averages (3, 7, and 14 days), volatility, day-to-day change, and seasonal encodings. The other twelve are the daily mean and maximum values of six key pollutants (PM$_{10}$, PM$_{2.5}$, CO, NO$_2$, SO$_2$, and O$_3$).

Spatial context is constructed using the five geographically nearest cities. Distances are calculated from latitude and longitude using the haversine formula. For a neighbour \(j\in\mathcal{N}_{5}(c)\), the normalized inverse-distance weight is
\begin{equation}
    \alpha_{cj,t}
    =
    \frac{
        \mathbb{I}_{j,t}/\max(d_{cj},1)
    }{
        \displaystyle
        \sum_{k\in\mathcal{N}_{5}(c)}
        \mathbb{I}_{k,t}/\max(d_{ck},1)
    },
    \label{eq:neighbor_weight}
\end{equation}
where \(d_{cj}\) is the distance in kilometres and \(\mathbb{I}_{j,t}\) indicates that the neighbour has a valid AQI value on day \(t\). The neighbourhood AQI and its dispersion are
\begin{equation}
    \bar{a}_{c,t}
    =
    \sum_{j\in\mathcal{N}_{5}(c)}
    \alpha_{cj,t}a_{j,t},
    \qquad
    s_{c,t}
    =
    \sqrt{
        \sum_{j\in\mathcal{N}_{5}(c)}
        \alpha_{cj,t}a_{j,t}^{2}
        -
        \bar{a}_{c,t}^{2}
    }.
    \label{eq:neighbor_context}
\end{equation}

The four neighbor features are \(\bar{a}_{c,t}\), \(s_{c,t}\), the local-to-neighbor gap \(a_{c,t}-\bar{a}_{c,t}\), and the neighborhood change \(\bar{a}_{c,t}-\bar{a}_{c,t-1}\). Only information available on day \(t\) is used. This construction should be understood as a fixed geographic neighborhood representation, not as a learned graph neural network.

\subsection{Dual-Head Forecasting and Warning}

The regression branch uses a histogram-based gradient-boosting model
\cite{friedman2001greedy} with all 30 features to predict the next-day maximum
AQI:
\begin{equation}
    \hat{y}_{c,t+1}
    =
    f_{\theta}\left(\mathbf{x}^{\mathrm{reg}}_{c,t}\right).
    \label{eq:regression_head}
\end{equation}

It minimizes duplicate-weighted absolute error and uses 180 iterations, a
learning rate of 0.08, 31 leaf nodes, 30 samples per leaf, and \(L_2\)
regularization of 1.0.

A separate classifier estimates the probability of Unhealthy-or-worse air:
\begin{equation}
    \hat{p}_{c,t+1}
    =
    \Pr\left(
        y_{c,t+1}\geq151
        \mid
        \mathbf{x}^{\mathrm{warn}}_{c,t}
    \right).
    \label{eq:warning_probability}
\end{equation}

This branch uses only the 24 temporal and pollutant features, and does not include location and neighbor information. Training applies both trajectory-group weights and inverse-frequency class weights. The warning threshold is chosen on the 2024 calibration set by maximizing duplicate-balanced F1 and then fixed for the 2025 test set.

\subsection{Risk-Conditional Conformal Calibration}

Prediction intervals are calibrated using only the 2024 residuals. Repeated
observations from the same trajectory group and target date are first collapsed
into one group-day. For absolute residuals \(r_i=|y_i-\hat{y}_i|\), the global
90\% conformal radius is
\begin{equation}
    q_{\alpha}
    =
    \operatorname{Quantile}_{\lceil(n+1)(1-\alpha)\rceil/n}
    (r_1,\ldots,r_n),
    \qquad \alpha=0.10.
    \label{eq:conformal_quantile}
\end{equation}

Because uncertainty changes with AQI severity, predictions are divided into four
predefined ranges: \(\hat{y}\leq100\), \(101\!-\!150\), \(151\!-\!200\), and
\(\hat{y}\geq201\). Separate lower and upper residual quantiles are estimated for
each range, giving
\begin{equation}
    \mathcal{C}_{c,t+1}
    =
    \left[
        \max(0,\hat{y}_{c,t+1}-q_b^-),
        \hat{y}_{c,t+1}+q_b^+
    \right].
    \label{eq:risk_conditional_interval}
\end{equation}

For AQI ranges with too few calibration examples (fewer than 100), we use a general interval instead of a range-specific one. We also add a safety mechanism that widens the forecast interval for high-risk predictions to ensure we catch at least 95\% of very unhealthy days (AQI 201+). All these settings are finalized before the 2025 evaluation.

Related features are grouped together for model auditing. Grouped permutation importance measures how each feature group affects test MAE. Local occlusion measures how replacing a feature group with its median changes the warning probability. These tools help explain the model, but do not prove causation.

\section{Experimental Protocol}
\label{sec:experimental_protocol}

\subsection{Chronological Evaluation Design}

Samples were split by target date rather than randomly in order to reflect real-world use. The model was trained on 2022--2023 data, calibrated on 2024, and tested on the
untouched 2025 period. August 19, 2022 was the first usable target date because 14 prior days were needed to construct the longest rolling features.

\begin{table}[t]
\centering
\caption{Chronological partition used in all experiments.}
\label{tab:data_split}
\small
\setlength{\tabcolsep}{4pt}
\begin{tabular}{lccc}
\hline
\textbf{Partition} & \textbf{Target dates} & \textbf{City-days} &
\textbf{Groups} \\
\hline
Training &
Aug.\ 19, 2022--Dec.\ 31, 2023 &
51,500 & 61 \\
Calibration &
Jan.\ 1--Dec.\ 31, 2024 &
37,698 & 61 \\
Future test &
Jan.\ 1--Nov.\ 23, 2025 &
33,681 & 61 \\
\hline
\end{tabular}
\end{table}

All models used the same valid samples, and city identity was not included as a feature. Standardization was fitted only on the training set. Trajectory weights from (\ref{eq:trajectory_weight}) prevented duplicated city trajectories from receiving extra influence. No 2025 outcomes were used for model selection, calibration, or warning-threshold tuning.

Transfer was also tested by holding out complete trajectory groups. One experiment used randomly selected groups, while the spatial experiment held out one of five geographic clusters. Both used the same number of groups, and all cities sharing an identical trajectory were kept together.

\subsection{Baselines and Ablation Settings}

The AQI regressor was compared with persistence, weekly-naive, ridge, Extra Trees, and XGBoost baselines. Persistence reused the maximum AQI from day \(t\), and weekly-naive used the value from seven days earlier. These provide important references because day-ahead AQI is strongly autocorrelated. The machine-learning baselines used the same 24 temporal and pollutant features and duplicate-balanced training weights.

To examine each feature group, the same histogram-gradient regressor was trained
with increasingly complete inputs:

\begin{enumerate}
    \item temporal and seasonal features;
    \item temporal and pollutant features;
    \item temporal, pollutant, and coordinate features; and
    \item all features, including neighbor context.
\end{enumerate}

This design isolates the added value of pollutants, location, and neighboring AQI conditions without changing the model.

The warning head was compared with current-AQI persistence, thresholded regression forecasts, and class-balanced logistic regression. Its decision threshold was selected on 2024 by maximizing duplicate-balanced \(F_1\) over values from 0.05 to 0.95, then frozen for the 2025 test.

Uncertainty experiments compared a global conformal interval, risk-conditional intervals, and the full severity-guarded method. Coverage auditing and trajectory weighting were retained throughout all experiments. Instead of just ablating a model feature, their removal would alter the valid data regime and unit of evidence.

\subsection{Evaluation Metrics and Statistical Analysis}

Day-ahead forecasts were evaluated using duplicate-balanced MAE, RMSE, and
\(R^2\), with MAE as the primary metric:
\begin{equation}
    \mathrm{MAE}
    =
    \frac{\sum_{i=1}^{N} w_i |y_i-\hat{y}_i|}
         {\sum_{i=1}^{N} w_i},
    \label{eq:weighted_mae}
\end{equation}
where \(w_i\) is the inverse size of the corresponding trajectory group. Warning
performance was measured using precision, recall, \(F_1\), balanced accuracy,
and AUROC. AUPRC and Brier score were reported as supplementary measures.
Prediction intervals were assessed by duplicate-balanced coverage and mean and
median width, both overall and across AQI-severity levels.

Statistical analysis was performed across the 61 unique trajectory groups.
Daily errors were first averaged within each group, and improvement over
persistence was calculated as
\begin{equation}
    \Delta_g
    =
    \mathrm{MAE}^{\mathrm{pers}}_g
    -
    \mathrm{MAE}^{\mathrm{CAFE}}_g,
    \label{eq:group_improvement}
\end{equation}
where \(\Delta_g>0\) favours CAFE-AQI. A 95\% confidence interval was estimated
using 2,000 group-level bootstrap samples, together with a one-sided Wilcoxon
signed-rank test. Resampling by trajectory group avoids treating duplicated and
temporally related observations as independent. All experiments used a fixed
random seed of 42.

\section{Results and Discussion}
\label{sec:results}

\subsection{Future-Year Forecasting Performance}

On the untouched 2025 test set, CAFE-AQI achieved an MAE of 9.76, an RMSE of 15.04, and an \(R^2\) of 0.910 (Table~\ref{tab:regression_results}). It reduced MAE by 27.1\% compared with persistence and by 66.5\% compared with the weekly-naive baseline. It also outperformed GPU-XGBoost and Extra Trees by 5.2\% and 6.7\%, respectively.

\begin{table}[t]
\centering
\caption{Duplicate-balanced results on the untouched 2025 test set. Runtime includes training and test inference.}
\label{tab:regression_results}
\scriptsize
\setlength{\tabcolsep}{2.8pt}
\begin{tabular}{l c c c c c c c}
\toprule
\textbf{Model} & \textbf{Feat.} & \textbf{Device} & \textbf{MAE} & \textbf{RMSE} & \(\boldsymbol{R^2}\) & \textbf{U-recall} & \textbf{Runtime (s)} \\
\midrule
HGB: temporal only & 12 & CPU & 12.52 & 17.64 & 0.877 & 0.814 & 4.22 \\
HGB: \(+\) pollutants & 24 & CPU & 9.95 & 15.31 & 0.907 & 0.815 & 3.81 \\
HGB: \(+\) coordinates & 26 & CPU & 9.90 & 15.16 & 0.909 & 0.815 & 3.73 \\
\textbf{CAFE-AQI:} & 30 & CPU & \textbf{9.76} & \textbf{15.04} & \textbf{0.910} & 0.817 & 4.04 \\
\textbf{\(+\) neigh. context} & & & & & & & \\
\midrule
GPU-XGBoost & 24 & GPU & 10.29 & 15.28 & 0.907 & 0.785 & 3.53 \\
Extra Trees & 24 & CPU & 10.46 & 15.30 & 0.907 & 0.754 & 6.61 \\
Ridge regression & 24 & CPU & 15.29 & 19.89 & 0.843 & 0.504 & 0.06 \\
Persistence & -- & -- & 13.39 & 19.26 & 0.853 & 0.868 & \(<0.01\) \\
Weekly naive & -- & -- & 29.12 & 38.79 & 0.403 & 0.709 & \(<0.01\) \\
\bottomrule
\end{tabular}
\end{table}

The ablation shows that pollutant measurements provided most of the gain, reducing MAE from 12.52 to 9.95. Adding coordinates and neighbor context produced smaller improvements (Section~5.5).

\begin{figure}[t]
    \centering
    \includegraphics[width=0.70\columnwidth]{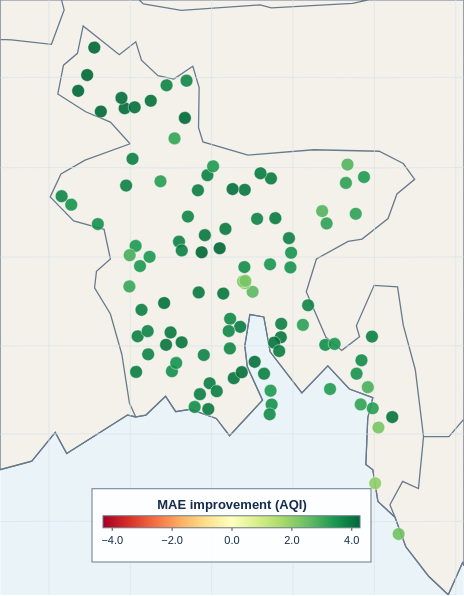}
    \caption{City-level 2025 MAE improvement over persistence. Marker size represents the number of 2025 test city-days. Green markers indicate lower MAE than persistence.}
    \label{fig:city_improvement}
\end{figure}

\begin{figure}[t]
    \centering
    \includegraphics[width=\columnwidth]{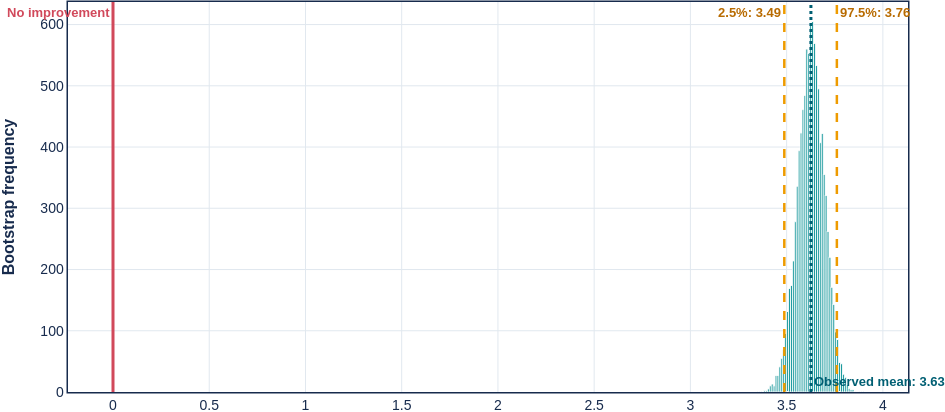}
    \caption{Bootstrap distribution of group-level improvement.}
    \label{fig:bootstrap_gain}
\end{figure}

CAFE-AQI also outperformed persistence across all 61 trajectory groups
(Figs.~\ref{fig:city_improvement} and~\ref{fig:bootstrap_gain}). The mean improvement was 3.63 AQI, with a group-bootstrap 95\% confidence interval of 3.49--3.76 AQI. The one-sided Wilcoxon test confirmed this improvement
(\(W=1891\), \(p=5.56\times10^{-12}\)).

By contrast, an invalid random city-day split produced an MAE of 6.15,
37.0\% below the valid future-year result. This shows how random splitting
can substantially overstate performance when related temporal and duplicated
trajectories appear in both training and test data.

\subsection{Unhealthy-Air Warning Performance}

Using the threshold of 0.475 selected on the 2024 calibration set, the CAFE-AQI
warning head achieved an \(F_1\)-score of 0.908, with 0.943 precision and 0.876
recall on the 2025 test set (Table~\ref{tab:warning_results}). It also obtained
an AUROC of 0.982 and an AUPRC of 0.976.

\begin{table}[t]
\centering
\caption{Duplicate-balanced warning results on the 2025 test set. Learned thresholds were selected using only 2024 data.}
\label{tab:warning_results}
\scriptsize
\setlength{\tabcolsep}{4pt}
\begin{tabular}{l|c|c|c|c}
\toprule
\textbf{Metric} & \textbf{Persistence} & \textbf{Regression} & \textbf{Logistic} & \textbf{CAFE-AQI} \\
 & & \textbf{forecast} & \textbf{warning baseline} & \textbf{warning head} \\

\midrule
Precision & 0.868 & \textbf{0.979} & 0.934 & 0.943 \\
Recall & 0.868 & 0.805 & 0.871 & \textbf{0.876} \\
F1 & 0.868 & 0.883 & 0.901 & \textbf{0.908} \\
Bal. acc. & 0.894 & 0.897 & 0.917 & \textbf{0.922} \\
AUROC & -- & -- & 0.980 & \textbf{0.982} \\
AUPRC & -- & -- & 0.971 & \textbf{0.976} \\
Brier & -- & -- & 0.0524 & \textbf{0.0488} \\
Threshold & 151 AQI & 151 AQI & 0.530 & 0.475 \\
\bottomrule
\end{tabular}
\end{table}

Thresholding the AQI forecast gave the highest precision but missed 19.5\% of
Unhealthy-or-worse events. The dedicated warning head improved recall from
0.805 to 0.876 and \(F_1\) from 0.883 to 0.908. Its advantage over logistic
regression was smaller, suggesting that the warning signal was already well
captured by the causal features, although CAFE-AQI remained the best overall
model.

\subsection{Prediction-Interval Reliability}

Both the fixed and risk-conditional intervals fell slightly below the nominal
90\% coverage, reaching 88.6\% and 88.7\%, respectively (Table~\ref{tab:uncertainty_transfer}). However, risk conditioning reduced the mean interval width from 45.81 to 42.19 AQI. Adding the severe-risk guard raised coverage to 92.1\%, while selectively widening intervals for high-risk forecasts.

\begin{table}[t]
\centering
\caption{Uncertainty and trajectory-transfer results on the test set.}
\label{tab:uncertainty_transfer}

\textbf{(a) Conformal interval performance}\\[3pt]
\scriptsize
\setlength{\tabcolsep}{3pt}
\begin{tabular}{lccc}
\toprule
\textbf{Method} &
\textbf{Coverage} &
\textbf{Mean width} &
\textbf{Median width} \\
\midrule
Fixed split-conformal & 0.886 & 45.81 & 45.81 \\
Risk-conditional & 0.887 & \textbf{42.19} & \textbf{42.14} \\
\textbf{Risk-conditional + guard} & \textbf{0.921} & 52.28 & \textbf{42.14} \\
\bottomrule
\end{tabular}

\vspace{6pt}

\textbf{(b) Trajectory-group transfer performance}\\[3pt]
\scriptsize
\setlength{\tabcolsep}{2.5pt}
\begin{tabular}{lccccc}
\toprule
\textbf{Evaluation} &
\textbf{Groups} &
\textbf{MAE} &
\textbf{RMSE} &
\(\boldsymbol{R^2}\) &
\textbf{U-recall} \\
\midrule
Seen-group future test & 0 & 9.76 & 15.04 & 0.910 & 0.817 \\
Random unseen groups & 13 & 10.22 & 15.88 & 0.899 & 0.791 \\
Unseen spatial block & 13 & 10.66 & 15.55 & 0.887 & 0.846 \\
\bottomrule
\end{tabular}

\end{table}

\begin{figure}[t]
    \centering
    \includegraphics[width=\columnwidth]{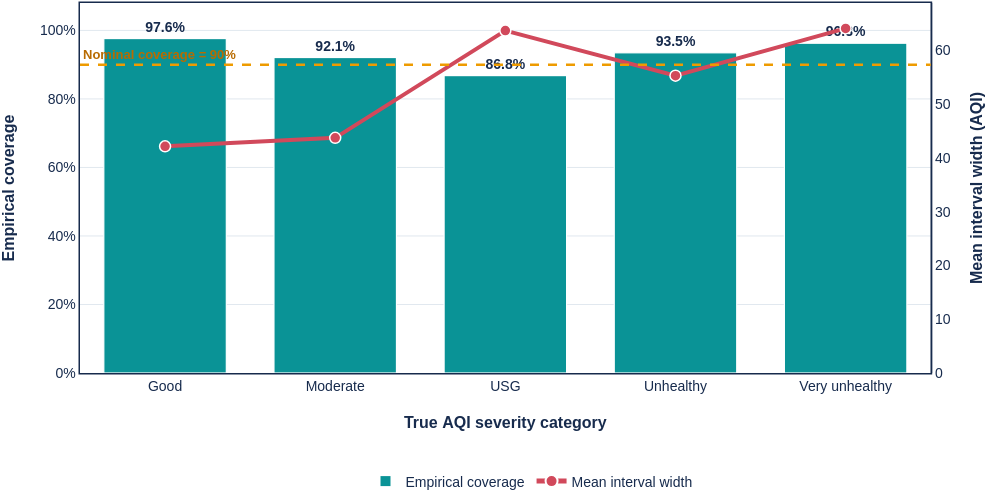}
    \caption{Coverage and interval width across true AQI categories.}
    \label{fig:severity_coverage}
\end{figure}

The guard detected 97.2\% of AQI 201-or-higher events while flagging 27.5\% of
the test observations. Coverage varied across severity levels and was lowest for Unhealthy for Sensitive Groups at 86.8\% (Fig.~\ref{fig:severity_coverage}). The interval results should therefore be interpreted as marginal coverage over the full test set, not as equal coverage within every AQI category.

\subsection{Transfer to Unseen Trajectory Groups}

When entire trajectory groups were held out, MAE rose from 9.76 to 10.22 for random unseen groups and to 10.66 for an unseen spatial block, increases of 4.7\% and 9.1\%, respectively. $R^2$ remained high at 0.899 and 0.887.

The larger drop under spatial transfer suggests that geographic extrapolation is harder than generalizing to randomly held-out groups. But the degradation was modest since the model retains useful structure for unseen locations within the same data source. These results should not be treated as external validation on a different monitoring network.

\subsection{Model-Audit Findings}

\begin{figure}[t]
    \centering
    \includegraphics[width=\columnwidth]{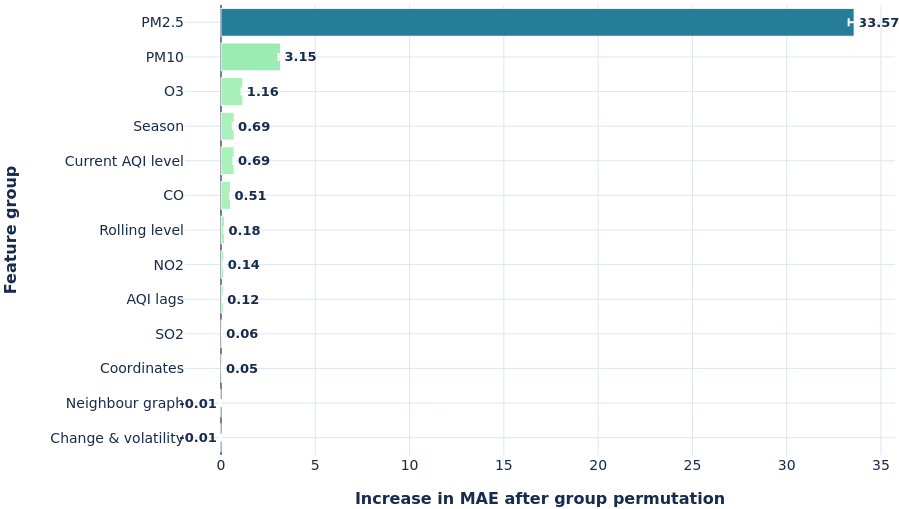}
    \caption{Grouped permutation importance of the regression features. PM$_{2.5}$ was the dominant predictive feature group.}
    \label{fig:grouped_importance}
\end{figure}

The audit showed that PM$_{2.5}$ drove most of the model's predictions. Permuting its features increased MAE by 33.57 AQI (Fig.~\ref{fig:grouped_importance}), compared with 3.15 for PM$_{10}$ and 1.16 for O$_3$. Coordinates and neighbor context had almost no effect (0.05 and \(-0.01\) AQI), consistent with their small ablation gain from 9.90 to 9.76 MAE. Thus, CAFE-AQI's performance mainly came from pollutant and temporal information rather than geographic context.

For Cox's B\=az\=ar on January 30, 2025, the model predicted a warning probability of 1.000. The observed maximum AQI was 172. Replacing PM$_{2.5}$ features with their training medians reduced this probability by 0.145. All other feature changes had negligible effects. This case shows the model's reasoning but does not imply causation.

\subsection{Overall Interpretation}

From the results, it is apparent that reliable next-day AQI forecasting did not require a large model. CAFE-AQI, which runs on a standard CPU, beat the persistence baseline for all 61 trajectory groups. The most important inputs were pollutant measurements, especially PM$_{2.5}$. Both neighbour data and dedicated warning head provided smaller gains. However, interval coverage remained uneven across severity levels, and so findings are limited to the audited Bangladesh data source.

\section{Limitations}
\label{sec:limitations}

This study has a few limitations. Identical AQI trajectories were found under multiple city labels; although trajectory-group weighting reduced their influence, the results should not be interpreted as evidence from 103 fully independent stations. Moreover, all experiments used the same Bangladesh AQI data source. The unseen-group and spatial-block tests show the model can generalize within this dataset, but independent validation at real monitoring stations is necessary before deployment. The model also does not include weather or emission data. The inclusion of these data could help predict sudden pollution events. Prediction intervals were not equally reliable across all AQI levels. The feature importance analysis explains what the model uses, but it does not explain what causes pollution. Future work should focus on independent observations, richer inputs, and continued calibration.

\section{Conclusion}
\label{sec:conclusion}

This study presented CAFE-AQI, a compact framework for next-day AQI forecasting, unhealthy-air warning, and uncertainty estimation. Under the strict 2025 future-year evaluation, CAFE-AQI achieved an MAE of 9.76 AQI, improving upon persistence by 27.1\%. It also produced the best warning performance, with an F1-score of 0.908, while the severity-guarded prediction intervals achieved 92.1\% overall coverage. Importantly, the forecasting improvement was observed across all 61 independent trajectory groups.

The findings show that careful chronological evaluation, duplicate-aware weighting, and pollutant information (particularly PM$_{2.5}$) were more important than model complexity. The geographic neighbor context provided a small but noticeable improvement. Although CAFE-AQI provides a useful and transparent forecasting framework, it is crucial to remember that, before thinking about real-world deployment, validation using independently verified monitoring data would be required.

\FloatBarrier


\begin{thebibliography}{00}

\bibitem{who2021aqg}
World Health Organization, ``WHO Global Air Quality Guidelines: Particulate Matter (PM$_{2.5}$ and PM$_{10}$), Ozone, Nitrogen Dioxide, Sulfur Dioxide and Carbon Monoxide,'' WHO, Geneva, Switzerland, 2021. [Online]. Available: https://www.who.int/publications/i/item/9789240034228

\bibitem{worldbank2022breathing}
World Bank, ``Breathing Heavy: New Evidence on Air Pollution and Health in Bangladesh,'' World Bank, Washington, DC, USA, 2022. [Online]. Available: https://documents.worldbank.org/en/publication/documents-reports/documentdetail/099440011162223258

\bibitem{usepa2024aqi}
U.S. Environmental Protection Agency, ``Technical Assistance Document for the Reporting of Daily Air Quality---The Air Quality Index (AQI),'' Office of Air Quality Planning and Standards, Research Triangle Park, NC, USA, Rep. EPA-454/B-24-002, May 2024. [Online]. Available: https://nepis.epa.gov/Exe/ZyPURL.cgi?Dockey=P101AP0Q.TXT

\bibitem{yi2018deepair}
X. Yi, J. Zhang, Z. Wang, T. Li, and Y. Zheng, ``Deep distributed fusion network for air quality prediction,'' in \emph{Proc. 24th ACM SIGKDD Int. Conf. Knowl. Discov. Data Mining}, 2018, pp. 965--973, doi: 10.1145/3219819.3219822.

\bibitem{agbehadji2024review}
I. E. Agbehadji and I. C. Obagbuwa, ``Systematic review of machine learning and deep learning techniques for spatiotemporal air quality prediction,'' \emph{Atmosphere}, vol. 15, no. 11, Art. no. 1352, 2024, doi: 10.3390/atmos15111352.

\bibitem{cerqueira2020evaluation}
V. Cerqueira, L. Torgo, and I. Mozeti\v{c}, ``Evaluating time series forecasting models: An empirical study on performance estimation methods,'' \emph{Mach. Learn.}, vol. 109, no. 11, pp. 1997--2028, 2020, doi: 10.1007/s10994-020-05910-7.

\bibitem{roberts2017crossvalidation}
D. R. Roberts \emph{et al.}, ``Cross-validation strategies for data with temporal, spatial, hierarchical, or phylogenetic structure,'' \emph{Ecography}, vol. 40, no. 8, pp. 913--929, 2017, doi: 10.1111/ecog.02881.

\bibitem{angelopoulos2023conformal}
A. N. Angelopoulos and S. Bates, ``Conformal prediction: A gentle introduction,'' \emph{Found. Trends Mach. Learn.}, vol. 16, no. 4, pp. 494--591, 2023, doi: 10.1561/2200000101.

\bibitem{paul2026aqi}
T. Paul, ``Bangladesh Air Quality Index (AQI) Dataset (2000--2025): Historical Hourly Air Pollution Data Across 103 Cities,'' Mendeley Data, V1, 2026, doi: 10.17632/9j447cynb9.1.

\bibitem{islam2025twentyfive}
M. R. Islam, R. Jany, M. F. Islam, R. B. Zaman, M. Saifuddin, and S. Sultana, ``Twenty-five years of air quality in Bangladesh: Trends, seasonality, and spatial pollution regimes,'' \emph{arXiv:2608.11652}, 2025. [Online]. Available: https://doi.org/10.48550/arXiv.2608.11652

\bibitem{friedman2001greedy}
J. H. Friedman, ``Greedy function approximation: A gradient boosting machine,'' \emph{Ann. Statist.}, vol. 29, no. 5, pp. 1189--1232, 2001, doi: 10.1214/aos/1013203451.

\end{thebibliography}
\end{document}